\pdfoutput=1

\documentclass[11pt]{article}

\usepackage[]{ACL2023}

\usepackage{times}
\usepackage{latexsym}

\usepackage[T1]{fontenc}

\usepackage[utf8]{inputenc}

\usepackage{microtype}
\usepackage{graphicx}
\usepackage{multirow}
\usepackage{tikz-dependency}
\usepackage{diagbox}
\usepackage{array}
\usepackage{algorithm}
\usepackage{algorithmicx}
\usepackage{algpseudocode}
\usepackage{mathtools}
\usepackage{algorithm}
\usepackage{amsmath,amssymb}
\usepackage{multicol}
\usepackage{fdsymbol}
\usepackage{multirow}
\usepackage{mathtools}
\usepackage[normalem]{ulem}
\usepackage{hyperref}
\usepackage{xcolor}
\usepackage{subcaption}
\usepackage{booktabs}

\usepackage{comment}
\usepackage{todonotes}
\usepackage{color,soul}

\usepackage{inconsolata}

\newcommand*{\NeuroX}{\textit{NeuroX}}

\title{NeuroX Library for Neuron Analysis of Deep NLP Models}

\author{Fahim Dalvi \\
  Qatar Computing Research \\
  Institute, HBKU \\
  \texttt{faimaduddin@hbku.edu.qa} \\\And
  Hassan Sajjad\thanks{{\hspace{1.5mm}The work was done while the author was at QCRI.}} \\
  Faculty of Computer Science \\
  Dalhousie University \\
  \texttt{hsajjad@dal.ca} \\\And
  Nadir Durrani \\
  Qatar Computing Research \\
  Institute, HBKU \\
  \texttt{ndurrani@hbku.edu.qa}}

\begin{document}
\maketitle
\begin{abstract}

Neuron analysis provides insights into how knowledge is structured in representations and discovers the role of neurons in the network. In addition to developing an understanding of our models, neuron analysis enables various applications such as debiasing, domain adaptation and architectural search. We present \NeuroX{}, a comprehensive open-source toolkit to conduct neuron analysis of natural language processing models. It implements various interpretation methods under a unified API, and provides a framework for data processing and evaluation, thus making it easier for researchers and practitioners to perform neuron analysis. The Python toolkit is available at \url{https://www.github.com/fdalvi/NeuroX}.\footnote{Demo Video available here: \url{https://youtu.be/mLhs2YMx4u8}}





\end{abstract}

\section{Introduction}

Interpretation of deep learning models is an essential attribute of trustworthy AI. Researchers have proposed a diverse set of methods to interpret models and answered questions such as: what linguistic phenomena are learned within representations, and what are the salient neurons in the network. For instance, a large body of work analyzed the concepts learned within representations of pre-trained models~\cite{durrani-etal-2019-one,liu-etal-2019-linguistic,tenney-etal-2019-bert,rogers-etal-2020-primer} and showed the presence of core-linguistic knowledge in various parts of the network. Several researchers have carried out this interpretation at a fine-grained level of neurons e.g.
\citet{durrani-etal-2020-analyzing,torroba-hennigen-etal-2020-intrinsic,antverg2021pitfalls} highlighted salient neurons w.r.t any linguistic property in the model and \citet{shappely_NIPS2017_7062, kamDhere,janizek2020explaining}  identified a set of neurons responsible for a given prediction. At a broader level, these methods can be categorized into representation analysis, neuron analysis and feature attribution methods respectively. \newcite{neuronSurvey} provides a comprehensive survey of these methods. 
\begin{figure}[!t]
    \centering
    \includegraphics[width=0.7\linewidth]{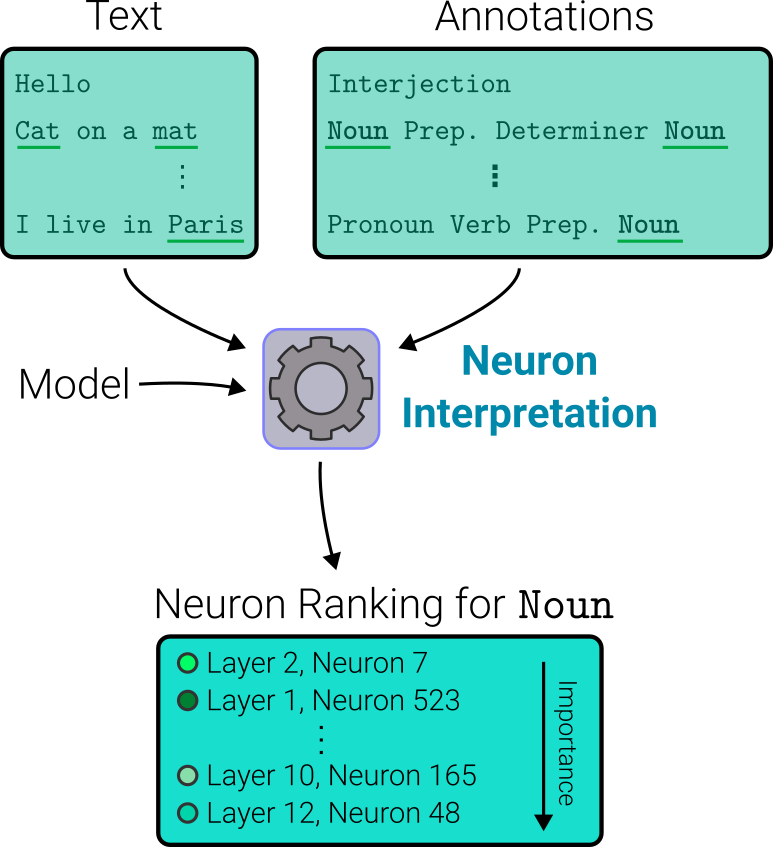}
    \caption{Simplified overview of Neuron Interpretation. Given an annotated text corpus, neuron interpretation methods aim to provide a ranking of neurons in a model w.r.t to their importance to one or more annotated properties (for e.g. "Noun" in this instance)}
    \label{fig:neuron-interpretation}
\end{figure}

A number of toolkits have been proposed to facilitate the interpretation of deep learning models. For instance, diagNNose~\cite{jumelet-2020-diagnnose} provides representation analysis and attribution methods. LIT~\cite{tenney-etal-2020-language-LIT} can be used to visualize attention and counterfactual explanations using feature attribution methods. Captum~\cite{kokhlikyan2020captum} integrates a large set of attribution methods under a consistent API.  All these tools facilitate the interpretability of models. However, due to the diverse ways of interpreting models, they do not cover all sets of methods. Specifically, most of the toolkits do not cover \textit{neuron interpretation} methods that discover and rank neurons with respect to a concept.

\begin{figure*}[!ht]
    \centering
    \includegraphics[width=0.95\linewidth]{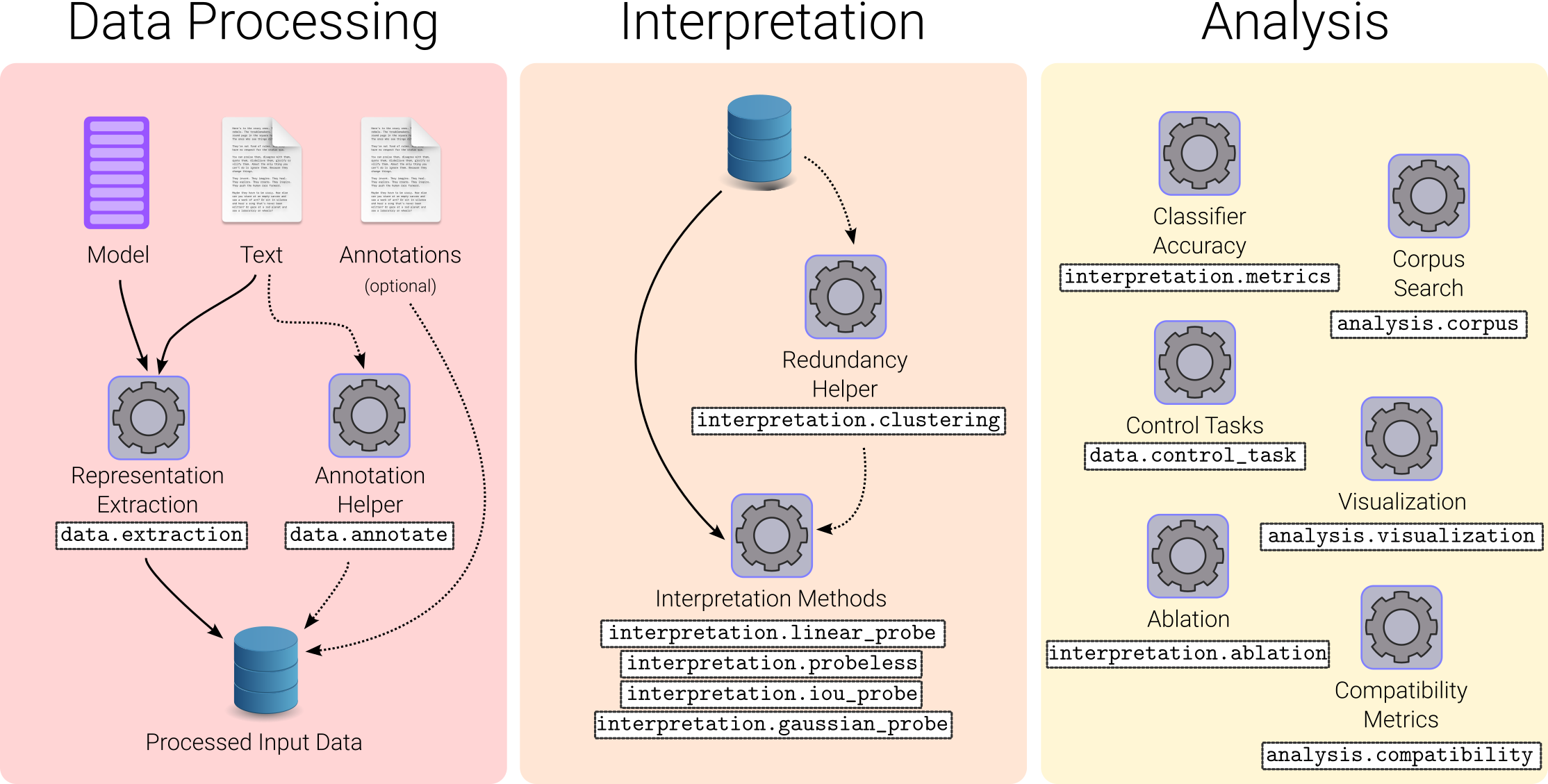}
    \caption{Overall design and architecture of the NeuroX toolkit, with references to their corresponding Python modules in the white boxes.}
    \label{fig:overview}
\end{figure*}

\textit{Neuron interpretation} analyzes and gives insight into how knowledge is structured within a representation. It discovers neurons with respect to a concept and provides a fine-grained interpretation of deep models. Figure \ref{fig:neuron-interpretation} provides a simplified high-level overview of how neuron interpretation methods operate. Given a model, some text and annotations,  these methods output a ranking of neurons with respect to their importance to one or more annotated concepts. The ability to interpret neurons enables applications such as debiasing models, controlling predictions of the models on the fly~\cite{bau2018identifying,suau2020finding}, neural architectural search~\cite{dalvi-2020-CCFS}, studying fine-tuned models \cite{durrani-etal-2021-transfer} and domain adaptation \cite{gu2021pruningthenexpanding}. To make neuron interpretation more accessible, we propose \NeuroX{}, an open-source Python toolkit to facilitate neuron interpretation of deep natural language processing (NLP) models. 

\NeuroX{} consists of three major components: i) data processing, ii) interpretation and iii) analysis. The data processing implements various ways to generate and upload data for analysis, extract activations and save them efficiently. The interpretation module implements six interpretation methods belonging to two different classes of methods. The analysis module brings together qualitative and quantitative methods to evaluate and visualize the discovered neurons. Figure~\ref{fig:overview} shows these components and how they interact with each other. We describe them in detail in the following sections. The toolkit itself is compatible with HuggingFace's transformers \cite{Wolf_Transformers_State-of-the-Art_Natural_2020} API and supports all transformer-based models.  

To the best of our knowledge, \NeuroX{} is the first toolkit that enables the interpretation of deep NLP models at the neuron level. It serves as a backbone to rapidly test new interpretation techniques using a unified framework and enables comparison and consistent evaluation of these techniques. The toolkit is easy to install and run:
\begin{verbatim}
    pip install neurox
\end{verbatim}
with detailed documentation is available at \url{https://neurox.qcri.org/docs}, including tutorials that showcase various capabilities of the toolkit to quickly get started with neuron interpretation.

\section{Data Processing}
\label{sec:dataprocessing}

\begin{table*}[!ht]
    \small
    \centering
    \begin{tabular}{c|c|c}
        \toprule
        Tokenizer & Input Sentence & Tokenized Sentence \\
        \midrule
        \texttt{bert-base-cased} & \texttt{"A good-looking house"} & \texttt{"[CLS] A good - looking house [SEP]"} \\
        \texttt{gpt2} & \texttt{"A good-looking house"} & \texttt{"A Ä good - looking Ä house"} \\
        \texttt{bert-base-cased} & \texttt{"Mauritians"} & \texttt{"[CLS] ma \#\#uri \#\#tian \#\#s [SEP]"} \\
        \texttt{gpt2} & \texttt{"Mauritians"} & \texttt{"M aur it ians"} \\
        \texttt{flaubert/flaubert\_base\_cased} & \texttt{"sport qu' on"} & \texttt{"<s> sport</w> qu</w> '</w> on</w> </s>"} \\
        \texttt{flaubert/flaubert\_base\_cased} & \texttt{"sport qu'"} & \texttt{"<s> sport</w> qu'</w> </s>"} \\
        \bottomrule
    \end{tabular}
    \caption{Tokenizers from different models tokenize the same input very differently, sometimes adding special characters at the first subword, or prefixing all subwords except the first subword etc. Sometimes the same model tokenizes the same word (qu') differently depending on the context.}
    \label{tab:tokenization}
\end{table*}

The \texttt{data} module is responsible for preparing all the inputs for the Interpretation and Analysis modules, as well as filtering the datasets to probe and interpret specific phenomena. As shown in Figure \ref{fig:overview}, the required inputs to the toolkit are: i) a model and ii) a text corpus annotated towards the property of interest (e.g. data annotated towards toxic word spans in the hate-speech-detection task). The interpretation module can then extract a neuron ranking, highlighting the saliency of the neurons in the model that capture this phenomenon. If annotations are not available, an annotation helper module is made available in the toolkit that can annotate tokens based on arbitrary phenomena e.g. suffixation, lexical properties, or using pre-existing vocabularies. Below we describe the various components of the \texttt{data} module in detail.

\subsection{Representation Extraction} 

Central to any neuron interpretation method are the neuron activations themselves, i.e. the magnitude of a neuron for any given input. While modern frameworks such as PyTorch and Tensorflow facilitate the extraction of intermediate neuron values for specific models via hooks, it is non-trivial to enable this generically, as the code to extract activations from specific network components (e.g. layers) is highly dependent on the underlying model implementation. \NeuroX{} implements generic extractors for specific popular frameworks and provides a highly-configurable PyTorch-based extractor.

\paragraph{Framework Specific Extractors} An example of a framework specific extractor is one for HuggingFace's \textit{transformers} models. The \textit{transformers} library exposes the intermediate output at each layer, which can then be used to access each neuron's (layer output) activation for any given input. 

\paragraph{Generic Extractors} Apart from framework specific extractors, the toolkit offers a generic extractor for PyTorch model, which runs as a two step process. In the first step, the toolkit maps out the architecture of the given model, and provides the user a \texttt{json} file that contains all the components of the model. The user can then choose exactly which of the components they need the activations for, which are then saved in the second step. 

\paragraph{Segmentation and De-Segmentation} A unique problem to text and NLP models is that of tokenization. For instance, every \textit{transformers} model has an associated \textit{tokenizer}, that breaks the tokens in an input sentence into subwords depending on the model's vocabulary. The same input can be tokenized differently by each model. To get a neuron's activation for a given input token regardless of tokenization, \NeuroX{} runs a detokenization procedure to combine the activation values on subwords into a single activation value. Table \ref{tab:tokenization} shows examples of how a sentence (and sometimes a word) can be tokenized differently depending on the underlying tokenizer and context. The toolkit also offers the user a choice on how the activation values across subwords should be combined such as  \texttt{first or last subword} or \texttt{average across subwords}.

\subsection{Annotation Helper}


While annotations are available for some linguistic properties, labeled data sets may not always be available. To carry out an interpretation in such a scenario, \NeuroX{} offers a helper module that can label the data with a positive or negative label per token. \texttt{data.annotate.annotate\_data} can annotate each token positively in three different ways:

\begin{enumerate}
    \item \textbf{Preset Vocabulary:} The 
    token exists in the given vocabulary.
    \item \textbf{Regular expression:} The 
    token matches with the given regular expression. For example, the expression \texttt{\^{}{\textbackslash}d+\$} 
    annotates all tokens that are composed of digits as positive samples.
    \item \textbf{Python function:} A function that returns a binary True/False for a given token. Arbitrary computation can be done inside this function. For instance, \texttt{lambda token: token.endswith("ing")} annotates all tokens ending with \textit{-ing} positively.
\end{enumerate}

\section{Interpretation Module}

\begin{table*}[ht]
    \small
    \centering
    \begin{tabular}{c|p{0.5\linewidth}|m{0.08\linewidth}|m{0.07\linewidth}|m{0.08\linewidth}}
        \toprule
         \shortstack{\textbf{Interpretation} \\ \textbf{Method}} & \textbf{Description} & \textbf{Supports Representation Analysis} & \textbf{Requires Training} & \textbf{Supports multi-class analysis} \\
         \midrule
         \texttt{Linear Probe} & Class of probing methods that use a linear classifier for neuron analysis. Specifically, the implementation provides probes introduced by 
         \begin{itemize}
             \item \newcite{Radford} (Classifier with L1 regularization)
             \item \newcite{lakretz-etal-2019-emergence} (Classifier with L2 regularization)
             \item \newcite{dalvi:2019:AAAI} (Classifier with Elastic Net regularization)
         \end{itemize} & Yes & Yes & Yes \\
         \texttt{Probeless} & A corpus-based neuron search method that obtains neuron rankings based on an accumulative strategy, introduced by \newcite{antverg2021pitfalls} & No & No & Yes \\
         \texttt{IoU Probe} & A mask-based method introduced by \newcite{Mu-Nips} that computes Intersection over Union between tokens representing a specific concept and tokens that have high activation values for specific neurons & No & No & No \\
         \texttt{Gaussian Probe} & A multivariate Gaussian based classifier introduced by \newcite{torroba-hennigen-etal-2020-intrinsic} that can probe for neurons whose activation values follow a gaussian distribution. & Yes & Yes & Yes \\
         \texttt{Mean Select} & A corpus-based neuron ranking method introduced by \newcite{fan2023evaluating} that derives neuron importances by looking at activation values across contexts where a concept appears. & No & No & Yes \\
         \bottomrule
    \end{tabular}
    \caption{An overview of the neuron interpretation methods currently implemented in the \NeuroX{} toolkit. 
    }
    \label{tab:probing-methods}
\end{table*}

The central module in the \NeuroX{} toolkit is the \texttt{interpretation} module, which provides implementations of several neuron and representation analysis methods. Table \ref{tab:probing-methods} shows a list of methods that are currently implemented in the toolkit, along with details of what each method's implementation supports.

The method implementations follow a consistent API to make it easy for the user to switch between them. Specifically, each method at least implements the following functions:

\begin{itemize}
    \item \texttt{method.train\_probe}: This function takes in the pre-processed data (extracted activations, prepared labels etc) as described in section \ref{sec:dataprocessing}, and returns back a probe that can be used to perform neuron analysis. Some methods do not require any training, in which case this function just stores the input for future use.
    \item \texttt{method.evaluate\_probe}: This function takes an evaluation set and returns the performance of the probe on the given set. The evaluation set itself can be a control task, and the output score can be computed using several pre-implemented metrics. Section \ref{sec:evaluation} discusses the various evaluation metrics in detail.
    \item \texttt{method.get\_neuron\_ordering}: This function returns an ordering/ranking of all the neurons being analyzed with respect to their importance to the task at hand. For instance, if the probe was trained to analyze \textit{Nouns}, this function will return a sorted list of neurons (by importance) that activate for Nouns in the given dataset.
\end{itemize}

The interpretation methods themselves may be able to probe multiple properties at the same time (multi-class probing), or only a single concept (binary probing). Additionally, some interpretation methods can also perform representation-level analysis, i.e. probe an entire layer rather than individual neurons. 

\paragraph{Redundancy Analysis:} \citet{dalvi-2020-CCFS} have shown that large neural networks learn knowledge redundantly where multiple neurons are optimized to activate on the same input. This is encouraged 
by the optimization choices such as dropouts which explicitly force neurons to learn in the absence of other neurons. In order to facilitate the analysis of redundant neurons, the toolkit provides a clustering based non-redundant neuron extraction method. Running the neurons through \texttt{interpretation.\-clustering.\-extract\_\-independent\_neurons} first before performing any probing can reduce the overall search space of neurons, and lead to better findings and analyses.

\section{Analysis and Evaluation}
\label{sec:evaluation}

The \texttt{analysis} module provides implementations of various evaluation and analysis methods. Some of these methods provide quantitative results like accuracy scores, while others allow users to perform qualitative analysis on neurons.


\subsection{Classifier Accuracy}
\label{sec:accuracy}

Classifier accuracy reciprocates the probing framework \cite{belinkov:2017:acl,hupkes2018visualisation}. Once a neuron ranking is obtained, a classifier is trained towards the task of interest (the intrinsic concept for which the probe was originally trained) with the selected neurons. The delta between oracle performance (accuracy using all the neurons) and the accuracy of the classifier using the selected neurons measures the efficacy of the ranking. 

\paragraph{Selectivity} It is important to ensure that the probe is truly representing the concepts encoded within the learned representations and not memorizing them during classifier training. \NeuroX{} enables control task selectivity, a measure proposed by \newcite{hewitt-liang-2019-designing} to mitigate memorization using the \texttt{data.control\_task} module.

\subsection{Ablation}
\label{sec:ablation}

An alternative approach used by \cite{dalvi:2019:AAAI} is to ablate all but the selected neurons in the trained probe. The \texttt{interpretation.ablation} allows manipulating the input data by keeping/filtering specific neurons in the order of their importance, allowing users to measure the drop in performance with selected neurons.

\subsection{Mutual Information}
\label{sec:MI}

Information theoretic metrics such as mutual information have also been used to interpret representations of deep NLP models~\cite{pimentel2020informationtheoretic}. Here, the goal is to measure the amount of information a representation provides about a linguistic concept. It is computed by calculating the mutual information between a subset of neurons and linguistic concepts.

\subsection{Compatibility Metrics}
\label{sec:voting}
Another set of evaluation metrics recently proposed by \citet{fan2023evaluating} carries out a pair-wise comparison of the discovered neurons across methods. While this strategy does not provide a direct evaluation of a neuron interpretation method, it provides an insight into how \textit{compatible} a method is with the other available methods. \NeuroX{} implements two compatibility metrics in the \texttt{analysis.compatibility} module: i) \texttt{Average Overlap} (which shows how aligned a method is with others) and ii) \texttt{NeuronVote} (which shows how well-endorsed the ranking of a method is by other methods).

\subsection{Qualitative Evaluation}
\label{sec:qe}

\begin{figure}[!ht]
	\centering
    \begin{subfigure}[b]{0.99\linewidth}
    \centering
    \includegraphics[width=\linewidth]{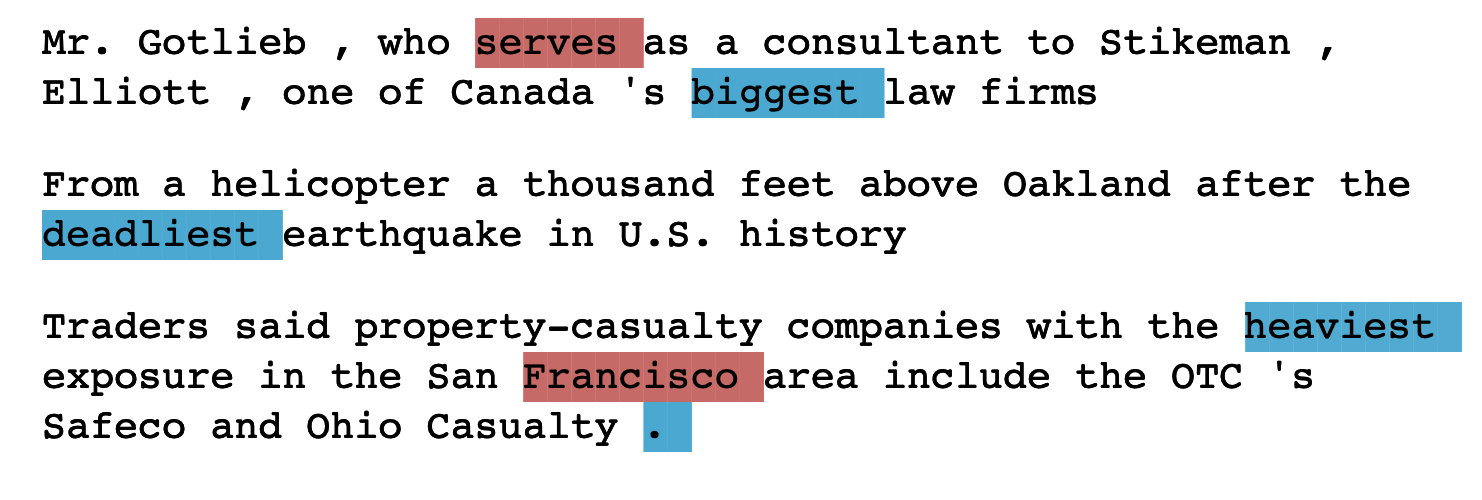}
    \caption{\label{fig:jjs} Superlative Adjective Neuron}
    \end{subfigure}
     \begin{subfigure}[b]{0.99\linewidth}
    \centering
    \includegraphics[width=\linewidth]{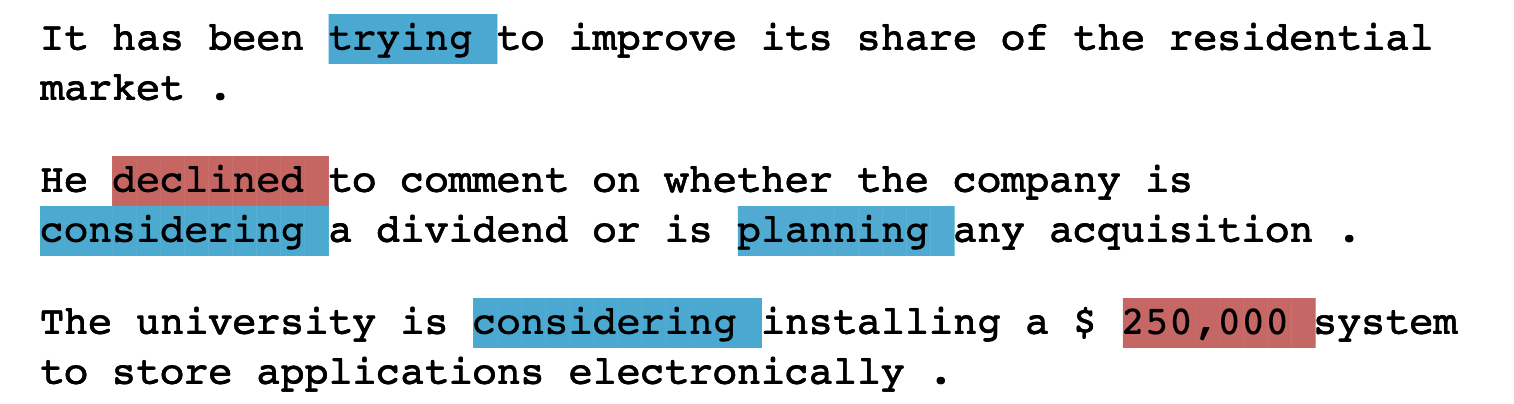}
    \caption{\label{fig:gerunds} Gerund Verb Neuron}
    \end{subfigure}
    \caption{\label{fig:vis} Visualizations (POS) -- Superlative Adjective and Gerund Verb Neurons}
\end{figure}

Visualizations have been used effectively to gain qualitative insights on analyzing neural networks \cite{karpathy2015visualizing,kadar-etal-2017-representation}. \NeuroX{} provides a text visualization module (\texttt{analysis.visualization}) that displays the activations of neurons w.r.t. to a concept (e.g. Figure~\ref{fig:vis}). The toolkit also allows corpus-based analysis in the \texttt{analysis.corpus} module by extracting the top $n$ words in a corpus that activate a neuron. Examples are shown in Table~\ref{tab:neuronExamples}. 

\begin{table*}[ht]

\centering					
    \small
    \begin{tabular}{l|l|l|l}									
    \toprule									
\textbf{Neuron}   & \textbf{concept} & \textbf{Model} & \textbf{Top-5 words} \\		
\midrule
     Layer 9: 624 & VBD & RoBERTa & supplied, deposited, supervised, paled, summoned \\
     Layer 2: 750 & VBG & RoBERTa & exciting, turning, seeing, owning, bonuses \\
     Layer 0: 249 & VBG & BERT & requiring, eliminating, creates, citing, happening \\
     Layer 1: 585 & VBZ & XLNet & achieves, drops, installments, steps, lapses, refunds \\
     Layer 2: 254 & CD & RoBERTa & 23, 28, 7.567, 56, 43 \\
     Layer 5: 618 & CD & BERT & 360, 370, 712, 14.24, 550 \\
     Layer 1: 557 & LOC & XLNet & Minneapolis, Polonnaruwa, Mwangura, Anuradhapura, Kobe \\
     Layer 5: 343 & ORG & RoBERTa & DIA, Horobets, Al-Anbar, IBRD, GSPC \\
     Layer 10: 61 & PER & RoBERTa & Grassley, Cornwall, Dalai, Bernanke, Mr.Yushchenko \\
     Layer 6: 132 & PER & BERT & Nick, Manie, Troy, Sam, Leith \\
     Layer 2: 343 & YOC & BERT &  1897, 1918, 1901, 1920,  Alam \\
    \bottomrule
    \end{tabular}
    \caption{Ranked list of words for some individual neurons, VBD: Past-tense verb, VBG: Gerund Verb, VBZ: Third person singular, CD: Numbers, LOC: Location, ORG: Organization, PER: Person, YOC: Year of the century}

\label{tab:neuronExamples}						
\end{table*}

\section{Miscellaneous Functions}
\subsection{Scalability}

Extracting, saving and working with neuron activations over large datasets and models can be very expensive, since each neuron's activation is saved for each token in the input corpus. To enable both disk- and runtime-savings, \NeuroX{} provides a low precision mode where all the activations are saved using 16-bit precision instead of the default 32/64-bit precision. This results in considerable storage/memory savings and also improves training/inference performance depending on the method and underlying hardware. The precision can be controlled by supplying the \texttt{dtype=float16} option to the extraction/interpretation methods.

\subsection{Disk Formats for Representations}
The toolkit offers flexibility to the user over the format used to save the neuron activations. Specifically, it offers readers and writers for a text-based format (\texttt{json}) and a binary format (\texttt{hdf5}). The binary format provides faster saving/loading performance, speeding up experiments with a large number of neurons or a large amount of text. On the other hand, the text-based format is considerably easier to debug.


\section{Related Work}

A number of toolkits have been made available to carry out the analysis and interpretation of neural network models. The What-If tool~\cite{wexler2019if} inspects machine learning models and provides users an insight into the trained model based on the predictions. Seq2Seq-Vis~\cite{strobelt-etal-2018-debugging} enables the user to trace back the prediction decisions to the input in neural machine translation models. Captum~\cite{kokhlikyan2020captum} provides generic implementations of a number of gradient and perturbation-based attribution algorithms. LIT~\cite{tenney-etal-2020-language-LIT} implements various methods of counterfactual explanations, attribution methods and visualization of attention. diagNNose~\cite{jumelet-2020-diagnnose} integrates representation analysis methods and attribution methods and finally, iModelsX~\cite{singh2023embgam} aims to provide natural explanations for datasets, which can provide insights into the models that are trained with these datasets. While these tools cover a number of interpretation methods, none of them facilitate neuron-level interpretation of NLP models. The LM-Debugger toolkit \cite{geva-etal-2022-lm} is an interactive debugger for transformer LMs, which provides a fine-grained interpretation and a powerful framework for intervening in LM behavior.

Ecco~\cite{alammar-2021-ecco} is a visualization based library that implements saliency methods and additionally enables visualization of neurons of the network. Similar to Ecco, the \NeuroX{} toolkit enables visualization of neurons of the network. In addition, we implement a wide range of neuron interpretation methods that can be accessed using a uniform API and provide various analysis and evaluation methods. Our toolkit empowers researchers to focus on specific parts of the neuron interpretation research such as interpretation, comparison or evaluation without worrying about setting up the rest of the pipeline like data processing, embedding extraction, integration with various pre-trained models, and evaluation of the method. NeuroX powers other interpretation analysis frameworks such as ConceptX \cite{alam2022conceptx} and NxPlain \cite{dalvi-etal-2023-nxplain}.

The previous version of NeuroX~\cite{dalvi2019neurox} only supported a specific machine translation library and one neuron interpretation method~\cite{durrani2022linguistic} as a GUI app. The current Python toolkit is a redesigned version with a unified architecture. It includes multiple features like a data processing module, numerous neuron interpretation and evaluation methods, and seamless integration with popular toolkits such as HuggingFace's \textit{transformers}.


\section{Conclusion and Future Work}

We presented \NeuroX{}, an open-source toolkit to carry out neuron-level interpretation of representations learned in deep NLP models. It maintains implementations of several neuron analysis methods under a consistent API, and provides implementations for preparing the data, analyzing neurons and evaluating the methods. In the future, \NeuroX{} plans to expand its extraction module to other frameworks like FairSeq and OpenNMT-py. In addition, we plan to integrate attribution based neuron saliency methods to add another class of interpretation methods to the toolkit.

\section{Acknowledgements}
We are grateful to all NeuroX contributors and users who have provided bug reports to improve the toolkit. Specifically, we would like to thank Ahmed Abdelali for testing and reporting bugs with the extraction implementations, David Arps for scalability improvements, Yimin Fan for implementing several interpretation methods and Yifan Zhang for working on detailed documentation and tutorials.

\section{Ethical Considerations}
The NeuroX toolkit provides a post hoc interpretation of pre-trained models. The toolkit makes a contribution towards improving the transparency of deep models and may discover biases present in these models. We do not foresee any direct ethical issues with respect to the developed toolkit. In terms of the neuron interpretation methods, the majority of them are based on the correlation between neurons and the input. One potential issue with such an interpretation is its faithfulness with respect to the knowledge used by the model in making predictions. However, this is not a limitation of the toolkit but a limitation of the research methods in general.

\bibliography{acl2023}
\bibliographystyle{acl_natbib}




\end{document}